\documentclass[final]{cvpr}

\usepackage[utf8]{inputenc} 
\usepackage[T1]{fontenc}    
\usepackage{url}            
\usepackage{booktabs}       
\usepackage{amsfonts}       
\usepackage{nicefrac}       
\usepackage{microtype}      
\usepackage{times}
\usepackage{epsfig}
\usepackage{graphicx}
\usepackage{amsmath}
\usepackage{amssymb}
\usepackage{subfiles}
\usepackage{multirow}
\usepackage{booktabs}
\usepackage{xcolor}
\usepackage{subcaption}
\usepackage{amssymb}
\usepackage{pifont}
\makeatletter
\@namedef{ver@everyshi.sty}{}
\makeatother
\usepackage{tikz}
\newcommand*\circled[1]{\tikz[baseline=(char.base)]{
            \node[shape=circle,draw,inner sep=0.4pt] (char) {#1};}}

\newcommand{\cmark}{{\color{blue}\ding{51}}}
\newcommand{\xmark}{{\color{red}\ding{55}}}

\usepackage[pagebackref=true,breaklinks=true,colorlinks,bookmarks=false]{hyperref}

\def\cvprPaperID{6452} 
\def\confYear{CVPR 2021}

\begin{document}

\title{SOON: Scenario Oriented Object Navigation with Graph-based Exploration}

\author{Fengda Zhu$^1$\hspace{6mm}
Xiwen Liang$^2$\hspace{6mm}
Yi Zhu$^3$\hspace{6mm}
Qizhi Yu$^4$ \hspace{6mm}
Xiaojun Chang$^1$\thanks{Corresponding author.} \hspace{6mm}
Xiaodan Liang$^2$ \\
%
$^1$Monash University \hspace{2mm} $^2$ Sun Yat-sen University \\ $^3$ University of Chinese Academy of Sciences \hspace{2mm} $^4$Zhijiang Laboratory\\
{\tt\small fengda.zhu@monash.edu \hspace{2mm} liangcici5@gmail.com \hspace{2mm} zhu.yee@outlook.com } \\
{\tt\small qyu@ieee.org \hspace{2mm} cxj273@gmail.com \hspace{2mm} xdliang328@gmail.com }
}

\maketitle
\pagestyle{empty}
\thispagestyle{empty}

\begin{abstract}
   The ability to navigate like a human towards a language-guided target from anywhere in a 3D embodied environment is one of the `holy grail' goals of intelligent robots. Most visual navigation benchmarks, however, focus on navigating toward a target from a fixed starting point, guided by an elaborate set of instructions that depicts step-by-step. This approach deviates from real-world problems in which human-only describes what the object and its surrounding look like and asks the robot to start navigation from anywhere. Accordingly, in this paper, we introduce a Scenario Oriented Object Navigation (SOON) task. In this task, an agent is required to navigate from an arbitrary position in a 3D embodied environment to localize a target following a scene description. 
   To give a promising direction to solve this task, we propose a novel graph-based exploration (GBE) method, which models the navigation state as a graph and introduces a novel graph-based exploration approach to learn knowledge from the graph and stabilize training by learning sub-optimal trajectories. 
   We also propose a new large-scale benchmark named From Anywhere to Object (FAO) dataset. To avoid target ambiguity, the descriptions in FAO provide rich semantic scene information includes: object attribute, object relationship, region description, and nearby region description. Our experiments reveal that the proposed GBE outperforms various state-of-the-arts on both FAO and R2R datasets. And the ablation studies on FAO validates the quality of the dataset. 
\end{abstract}

\section{Introduction}
Recent research efforts~\cite{wu2018building, gupta2020cognitive, fried2018speaker, wang2018reinforced, mezghani2020learning, thomason2019vision, majumdar2020improving} have achieved  great success in embodied navigation tasks. 
The agent is able to reach the target by following a variety of instructions, such as a word (\emph{e.g.} object name or room name)~\cite{wu2018building, savva2017minos}, a question-answer pair~\cite{das2018embodied, gordon2018iqa}, a natural language sentence~\cite{anderson2018vision} or a dialogue consisting of multiple sentences~\cite{thomason2019vision,ZZLJCL20}. 

However, these navigation approaches are still far from real-world navigation activities. Current vision language based navigation tasks such as Vision-language Navigation (VLN)~\cite{anderson2018vision}, Navigation from Dialog History (NDH)~\cite{thomason2019vision}  focus on navigating to a target by a fixed trajectory, guided by an elaborate set of instructions that outlines every step. These approaches fail to consider the case in which the complex instruction provided only target description while the starting point is not fixed. In real-world applications, people often do not provide detailed step-by-step instructions and expect the robot to be capable of self-exploration and autonomous decision-making. 
We claim that the ability to navigate towards a language-guided target from anywhere in a 3D embodied environment like human would be of great importance to an intelligent robot. 

\begin{figure*}[t]
	\centering
    \includegraphics[width=0.90\linewidth]{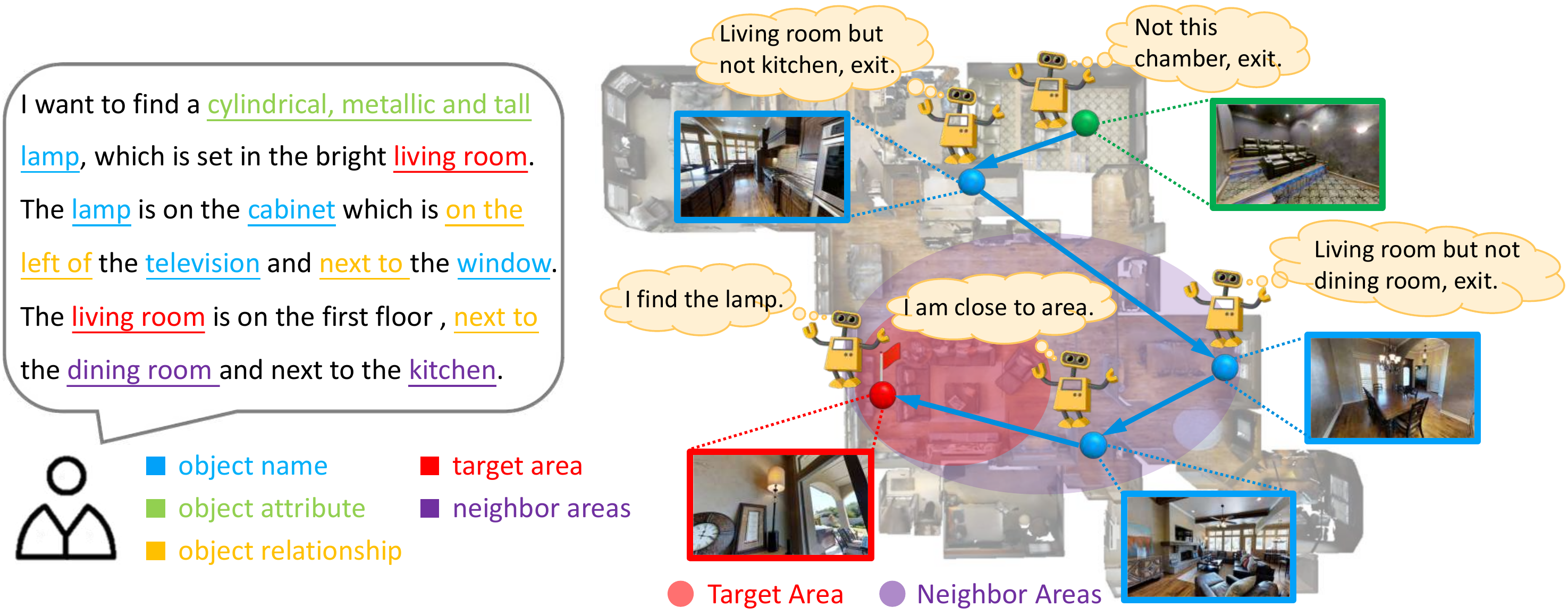}
	\caption{
	An example of the navigation process in SOON. An agent receives a complex natural language instruction consisting of multiple kinds of descriptions (left-hand side). During the agent navigates among different rooms, it searches a larger-scale area first, then gradually narrows down the search scope according to the visual scene and the instructions. }
	\label{fig:overview}
	\vspace{-8pt}
\end{figure*}

To address these problems, we propose a new task, named Vision Situated Object Navigation (SOON), where an agent is instructed to find a thoroughly described target object inside a house. 
The navigation instructions in SOON are target-oriented rather than step-by-step babysitter as in previous benchmarks. 
There are two major features that makes our task unique: target orienting and starting independence. 
A brief example of a navigation process in SOON is illustrated in Fig.~\ref{fig:overview}. 
Firstly, different from conventional object navigation tasks defined in~\cite{wu2018building, savva2017minos}, instructions in SOON play a guidance role in addition to distinguish a target object class. 
An instruction contains thorough descriptions to guide the agent to find a unique object from anywhere in the house. 
After receiving an instruction in SOON, the agent first searches a larger-scale area according to the region descriptions in the instruction, and then gradually narrows the search space to the target area. 
Compared with step-by-step navigation settings~\cite{anderson2018vision} or object-goal navigation settings~\cite{wu2018building}, this kind of coarse-to-fine navigation process is more closely resembles a real-world situation. 
Moreover, the SOON task is starting-independent. Since the language instructions contain geographic region descriptions rather than trajectory specific descriptions, 
they do not limit how the agent finds the target. 
By contrast, in step-by-step navigation tasks such as Vision Language Navigation~\cite{anderson2018vision} or Cooperative Vision-and-Dialog Navigation~\cite{thomason2019vision}, any deviation from the directed path may be considered as an error~\cite{ilharco2019general}. 
We present an overall comparison between the SOON task and existing embodied navigation tasks in Tab.~\ref{table:difference}. 

\begin{table*}
	\centering
	\resizebox{1.0\textwidth}{!}{
	\setlength{\tabcolsep}{0.6em}
    {\renewcommand{\arraystretch}{0.94}
	\begin{tabular}{l|ccc|cc|ccc}
		\hline
        \multirow{2}{*}{\textbf{Dataset}} & \multicolumn{3}{c|}{Instruction Context} 
        & \multicolumn{2}{c|}{Visual Context}
        & \multicolumn{1}{c}{Starting}
        & \multicolumn{1}{c}{Target}\\
        & Human & Content & Unamb. & Real-world & Temporal & Independent & Oriented \\ 
		\hline
		House3D~\cite{wu2018building} & \xmark & Room Name & \cmark & \xmark & Dynamic & \cmark & \xmark \\
		MINOS~\cite{savva2017minos} & \xmark & Ojbect Name & \cmark & \cmark & Dynamic & \cmark & \xmark \\
		
		EQA~\cite{das2018embodied}, IQA~\cite{gordon2018iqa} & \xmark & QA & \cmark & \xmark & Dynamic & \cmark & \xmark \\
		MARCO~\cite{macmahon2006walk}, DRIF~\cite{blukis2018mapping} & \cmark & Instruction & \cmark & \xmark & Dynamic & \xmark & \cmark\\
		R2R~\cite{anderson2018vision} & \cmark & Instruction & \cmark & \cmark & Dynamic & \xmark & \cmark\\
		TouchDown~\cite{chen2019touchdown} & \cmark & Instruction & \cmark & \cmark & Dynamic & \xmark & \cmark\\
		VLNA~\cite{nguyen2019vision}, HANNA~\cite{nguyen2019help} & \xmark & Dialog & \xmark & \cmark & Dynamic & \xmark & \cmark\\
		TtW~\cite{de2018talk} & \cmark & Dialog & \cmark & \cmark & Dynamic & \xmark & \cmark\\
		CVDN~\cite{thomason2019vision} & \cmark & Dialog & \xmark & \cmark & Dynamic & \xmark & \cmark\\
		REVERIE~\cite{qi2019reverie} & \cmark & Instruction & \cmark & \cmark & Dynamic & \xmark & \cmark\\
		\hline
		FAO (Ours) & \cmark & Instruction & \cmark & \cmark & Dynamic & \cmark & \cmark\\
		\hline
	\end{tabular} 
	}
	}
	\caption {Compared with existing datasets involving embodied vision and language tasks. }
	\label{table:difference}
	\vspace{-10pt}
\end{table*}

In this work, We propose a novel Graph-based Semantic Exploration (GBE) method to suggest a promising direction in approaching SOON. 
The proposed GBE has two advantages compared with previous navigation works~\cite{anderson2018vision, fried2018speaker, wang2018reinforced}. 
Firstly, GBE models the navigation process as a graph, which enables the navigation agent to obtain a comprehensive and structured understanding of observed information. It adopts graph action space to significantly merge the multiple actions in conventional sequence-to-sequence models~\cite{anderson2018vision,fried2018speaker,wang2018reinforced} into one-step decision. 
Merging actions reduces the number of predictions in a navigation process, which makes the model training more stable. 
Secondly, different from other graph-based navigation models~\cite{deng2020evolving, chaplot2020learning} that use either imitation learning or reinforcement to learn navigation policy,
the proposed GBE combines the two learning approaches and proposes a novel exploration approach to stabilize training by learning from sub-optimal trajectories. 
In imitation learning, the agent learns to navigate step by step under the supervision of ground truth label. 
It causes severe overfitting problem since labeled trajectories occupy only a small proportion of the large trajectory space. 
In reinforcement learning, the navigation agent explores large trajectory space, and learn to maximize the discounted reward. 
Reinforcement learning leverages sub-optimal trajectories to improve the generalizability. 
However, the reinforcement learning is not an end-to-end optimization method, which is difficult for the agent to converge and learn a robust policy. 
We propose to learn the optimal actions in trajectories sampled from imperfect GBE policy to stabilize training while exploration. Different from other RL exploration methods, the proposed exploration method is based on the semantic graph, which is dynamically built during the navigation. Thus it helps the agent to learn a robust policy while navigating based on a graph. 

To investigate the SOON task, we propose a large-scale From Anywhere to Object (FAO) benchmark. This benchmark is built on the Matterport3D simulator, which comprises 90 different housing environments with real image panoramas. FAO provides 4K sets of annotated instructions with 40K trajectories. As Fig.~\ref{fig:overview} (left) shows, one set of the instruction contains three sentences, including four levels of description: i) the color and shape of the object; ii) the surrounding objects along with the relationships between these objects and the target object; iii) the area in which the target object is located and the neighbour areas. Then, the average word number of the instructions is 38 (R2R is 26), and the average hop of the labeled trajectories is 9.6 (R2R is 6.0). Thus our dataset is more challenging than other tasks. 

We present experimental analyses on both R2R and FAO datasets to validate the performance of the proposed GBE and the quality of FAO dataset. 
The proposed GBE significantly outperforms previous previous VLN methods without pretraining or auxiliary tasks on R2R and SOON tasks. 
We further provide human performance on the test set of FAO to quantify the human-machine gap. 
Moreover, by ablating vision and language modals with different granularity, we validate that our FAO dataset contains rich information that enables the agent to successfully locate the target. 

\section{Related Work}
\noindent\textbf{Vision Language Navigation} 
Navigation with vision-language information has attracted widespread attention, since it is both widely applicable and challenging. 
Anderson \emph{et al.}~\cite{anderson2018vision} propose Room-to-Room (R2R) dataset, which is the first Vision-Language Navigation (VLN) benchmark combining real imagery~\cite{chang2017matterport3d} and natural language navigation instructions. 
In addition, the TOUCHDOWN dataset~\cite{chen2019touchdown} with natural language instructions is proposed for street navigation. 
To address the VLN task, Fried \emph{et al.} propose a speaker-follower framework~\cite{fried2018speaker} for data augmentation and reasoning in supervised learning, along with a concept named "panoramic action space" proposed to facilitate optimization. Wang \emph{et al.}~\cite{wang2018reinforced} demonstrate the benefit to combine imitation learning~\cite{bojarski2016end, ho2016generative} and reinforcement learning~\cite{mnih2016asynchronous, schulman2017proximal}. Other methods~\cite{wang2018look, ma2019self, ma2019the, tan2019learning, ke2019tactical, huang2019transferable} have been proposed to solve the VLN tasks from various angles. 
Inspired by the success of VLN, many datasets based on natural language instructions or dialogues have been proposed. VLNA~\cite{nguyen2019vision} and HANNA~\cite{nguyen2019help} are environments in which an agent receives assistance when it gets lost. TtW~\cite{de2018talk} and CVDN~\cite{thomason2019vision} provide dialogues created by communication between two people to reach the target position. Unlike the above methods, REVERIE~\cite{qi2019reverie} introduces a remote object localization task; in this task, an agent is required to find an object in another room that is unable to see at the beginning. 
The proposed SOON task is a coarse-to-fine navigation process, which navigates towards a target from anywhere following a complex scene description. 
An overall comparison between the SOON task and existing embodied navigation tasks is shown in Tab.~\ref{table:difference}.

\begin{figure*}[!t]
    \begin{minipage}{0.7\linewidth}
        \centering
        \includegraphics[width=\textwidth]{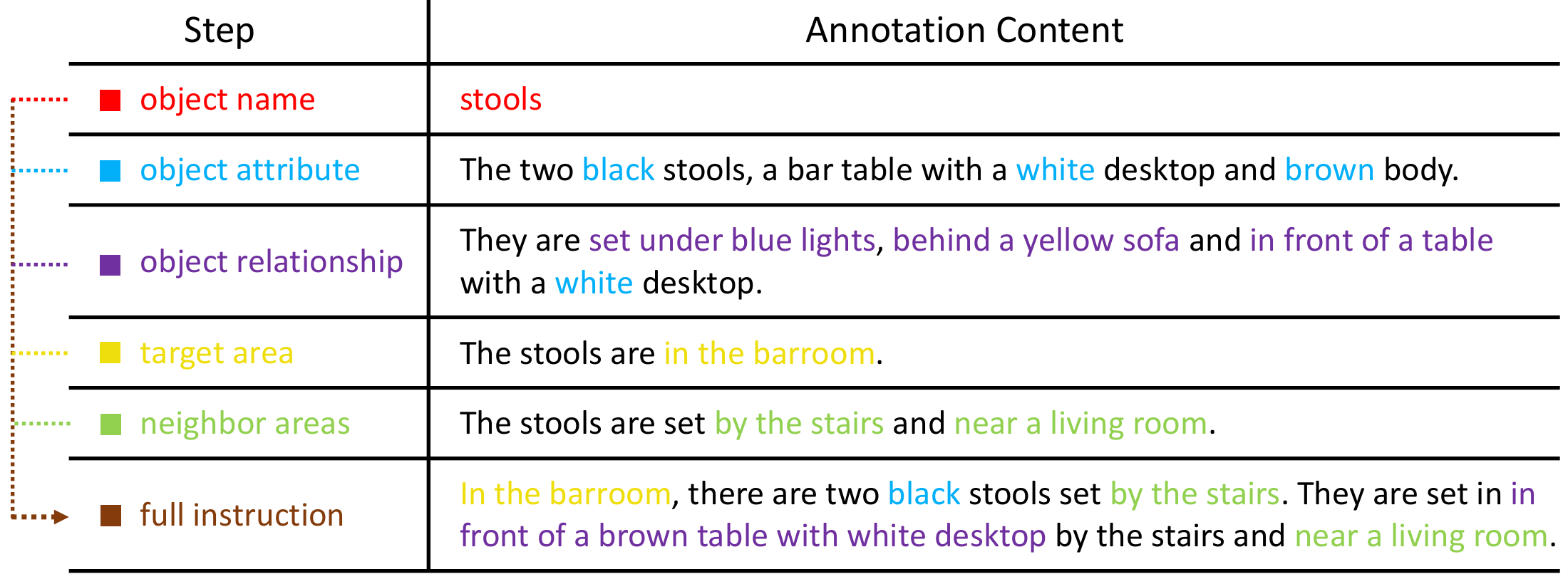}
        \vspace{-15pt}
        \caption{An example of annotating instructions in 6 steps.}
        \label{fig:instruction}
    \end{minipage} 
    \hfill
    \begin{minipage}{0.27\linewidth}
        \centering
        \includegraphics[width=\textwidth]{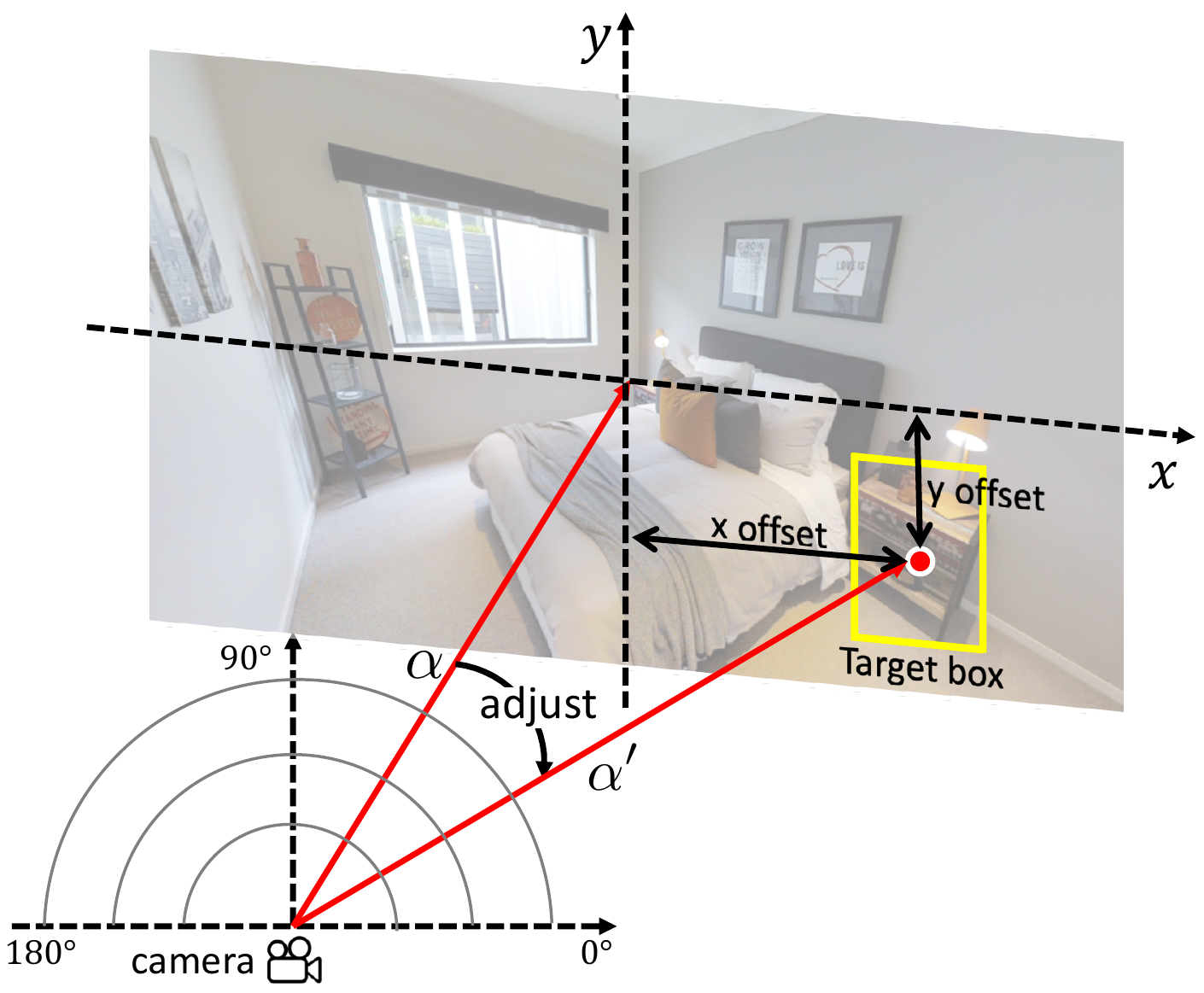}
        \caption{Converting a 2D bounding box into Polar coordinate. }
        \label{fig:ploar}
    \end{minipage}%
    \vspace{-10pt}
\end{figure*}

\noindent\textbf{Mapping and Planning} Classical SLAM-based methods~\cite{thrun2005probabilistic, davison1998mobile, gupta2020cognitive, fang2019scene, henriques2018mapnet, avraham2019empnet} build a 3D map with LIDAR, depth or structure, and then plan navigation routes based on this map. Due to the development of photo-realistic environments~\cite{anderson2018vision, chen2019touchdown, xia2018gibson} and efficient simulators~\cite{dosovitskiy2017carla, savva2017minos, savva2019habitat}, deep learning-based methods~\cite{mnih2015human, levine2016end, zhu2017target} have become feasible ways of training a navigation agent. 
Since deep learning methods have revealed their ability in feature engineering, end-to-end agents are becoming popular. 
Later works~\cite{fang2019scene, zhang2017neural, mezghani2020learning} adopt the idea of SLAM and introduce a memory mechanism, a method combining classical mapping methods and deep learning methods for generalization and long-trajectory navigation purposes. 
Recent works~\cite{chaplot2020learning, deng2020evolving, chaplot2020object} model the navigation semantics in graphs and achieve great success in embodied navigation tasks. 
Different from previous work~\cite{deng2020evolving} that only trains the agent using labeled trajectories by imitation learning, our works introduce reinforcement learning in policy learning and propose a novel exploration method to learn a robust policy. 

\section{Scenario Oriented Object Navigation}


\noindent\textbf{Task Definition of SOON} 
We propose a new Scenario Oriented Object Navigation (SOON) task, in which an agent navigates from an arbitrary position in a 3D embodied environment to localize a target object following an instruction. 
The task includes two sub-tasks: navigation and localization. 
We consider a navigation to be a \emph{success} if the agent navigates to a position close to the target (<3m);
and we consider the localization to be a \emph{success} if the agent correctly locates the target object in the panoramic view based on the success of navigation. 
To ensure that the target object can be found regardless of the agent's starting point, the instruction consists of several parts: i) object attribute, ii) object relationship, iii) area description, and vi) neighbor area descriptions. 
An example to demonstrate different parts of description is shown in Fig.~\ref{fig:instruction}. 
In step $t$ in navigation, the agent observes a panoramic view $v_t$, containing RGB and depth information. 
Meanwhile, the agent receives neighbour node observations $U_t=\{u^1_t,...,u^k_t\}$, which are the observations of $k$ reachable positions from the current position. 
All reachable positions in a house scan are discretized into a navigation graph, and the agent navigates between nodes in the graph. 
For each step, the agent takes an action $a$ to move from the current position to a neighbor node or stop. 
In addition to RGB-D sensor, the simulator provides a GPS sensor to inform the agent of its x, y coordinates. Also the simulator provides the indexes of the current node and candidate nodes. 


\noindent\textbf{Polar Representation} 
\label{sec:polar}
REVERIE~\cite{qi2019reverie} annotates 2D bounding boxes in 2D views to represent the location of objects. The 2D views are separated from the panoramic views of the embodied simulator. 
This way of labeling has two disadvantages: 1) some object separated by 2D views is not labeled; 2) 2D image distortion introduces labeling noise. 
We adopt the idea of \emph{Point Detection}~\cite{oquab2015object, zhu2017soft} and represent the location by polar coordinates, as shown in Fig.~\ref{fig:ploar}. 
First, we annotate the object bounding box with four vertices $\{ p_1, p_2, p_3, p_4 \}$. Then, we calculate the center point by $p_c$. After that, we convert the 2D coordinates into an angle difference between the original camera ray $\alpha$ and the adjusted camera ray $\alpha{}'$. 
\begin{figure*}[t]
\centering
\includegraphics[width=0.85\textwidth]{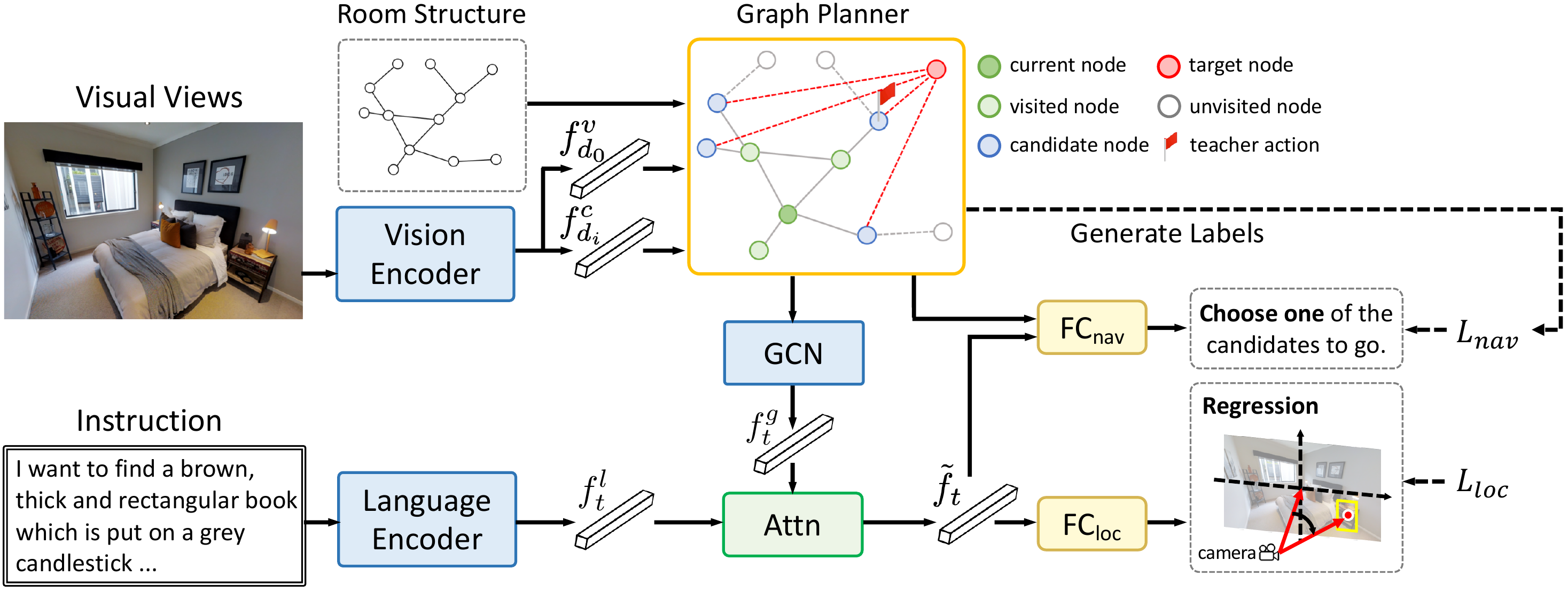}
\caption{
An overview of Graph-Based Semantic Exploration (GBE) model. 
Visual views are encoded by vision encoder and instructions are encoded by language encoder. The graph planner models the room semantics based on vision embeddings and the room structure information. 
GBE employs a GCN to embed graph nodes and output a graph embedding. 
Then, GBE outputs a cross-modal feature based on the graph embedding feature and language features. 
After that, GBE uses the cross-modal feature to predict the navigation action and regress the target location. 
}
 \vspace{-7pt}
\label{fig:model}
\end{figure*}
\section{Graph-based Semantic Exploration}
\label{sec:GBE}
We present the Graph-based Semantic Exploration (GBE) method in this section. 
The pipeline of the GBE is shown in Fig.~\ref{fig:model}. 
Our vision encoder $g$ and language encoder $h$ are built on a common practice of vision language navigation~\cite{wang2018reinforced, tan2019learning, zhu2019vision}. 
Subsequently, we introduce the graph planner in GBE, which models the structured semantics of visited places. 
Finally, we introduce our exploration method based on the graph planner. 

\noindent\textbf{Graph-based Navigation}
Memorizing viewed scenes and explicitly model the navigation environment are helpful for long-term navigation. 
Thus, we introduce a graph planner to memorize the observed features and model the explored areas as a feature graph. 
The graph planner maintains a node feature set $\mathcal{V}$, an edge set $\mathcal{E}$ and a node embedding set $\mathcal{M}$. 
The node feature set $\mathcal{V}$ is used to store node features and candidate features generated from visual encoder $g$. 
The edge set $\mathcal{E}$ dynamically updated to represent the explored navigation graph. 
The embedding set $\mathcal{M}$ stores the intermediate node embeddings, which are updated by GCN~\cite{kipf2016semi}. 
The node features in $\mathcal{M}$, noted as $f^{\mathcal{M}}_{n_i}$, are initialized by the feature of the same position in $\mathcal{V}$. 
At step $t$, the agent navigates to a position whose index is $d_0$, and receives a visual observation $v_t$ and the observations of neighbor nodes are $U_t=\{u^1_t,...,u^k_t\}$, 
where $k$ is the number of the neighbors and $N_t=\{n_1,...,n_k\}$ are node indexes of the neighbors. 
The visual observation and neighbor observations are embedded by the visual encoder $g$: 
\begin{equation}
\begin{split}
f^v_{n_0} &= g(v_t) \\
f^{c}_{n_i} &= g(u^i_t),
\end{split}
\end{equation}
where $n_0$ stands for the current node, and $n_i (1 \le i \le n)$ are the node it connects with. 
The graph planners add the $f^v_t$ and $f^{u,i}_t$ into $\mathcal{V}$: 
\begin{equation}
\mathcal{V} \leftarrow \mathcal{V} \cup  \{f^v_{n_0}, f^{u}_{n_1},...,f^{u}_{n_k}\}. 
\end{equation}
For an arbitrary node $n_i$ in the navigation graph, its node feature is represented by $\mathcal{V}$ following two rules: 
1) if a node $n_i$ is visited, its feature $f_{n_i}$ is represented by $f^{v}_{n_i}$; 
2) if a node $n_i$ is not visited but only observed, its feature is represented by $f^{u}_{n_i}$; 
3) since a navigable position is able to be observed from multiple different views, the unvisited node feature is represented by the average value of all observed features. 
The graph planner also updates the edge set $\mathcal{E}$ by:
\begin{equation}
\mathcal{E} \leftarrow \mathcal{E} \cup \{(n_0, n_1),(n_0, n_2),...,(n_0, n_k)\}. 
\end{equation}
An edge is represented by a tuple consists of two node indexes, indicating that two nodes are connected. 
Then, $\mathcal{M}$ is updated by GCN based on $\mathcal{V}$ and $\mathcal{E}$: 
\begin{equation}
\mathcal{M} \leftarrow \mathrm{GCN}(\mathcal{M}, \mathcal{E}). 
\end{equation}
To obtain comprehensive understanding of the current position and nearby scene, 
we define the output of the graph planner as: 
\begin{equation}
f^g_t =  \frac{1}{k+1} \sum_{i=0}^{k} f^{\mathcal{M}}_{n_i}, 
\end{equation}
$f^g_t$ and language feature $f_t^l$ perform cross-modal matching~\cite{wang2018reinforced} and output $\Tilde{f_t}$. 
GBE uses the $\Tilde{f_t}$ for two tasks: navigation action prediction and target object localization. 
The candidates to navigate are all observed but not visited nodes whose indexes are $C = \{c_1,...,c_{|C|}\}$, where $|C|$ is the number of candidates. 
The candidate feature are extracted from $\mathcal{V}$, denoted as $\{f_{c_1},...,f_{c_{|C|}}\}$. 
The agent generates a probability distribution $p_t$ over candidates for action prediction, and outputs regression results $\hat{l^h_i}$ and $\hat{l^e_i}$ standing for heading and elevation values for localization: 
\begin{equation}
\begin{split}
    z_i &= W_{nav} [\Tilde{f_t}, f_{c_i}], \\
    p_t(a_{c_i}) &= \mathrm{exp}(z_i) / \sum_j \mathrm{exp}(z_j), \\
    [\hat{l^h_i}, \hat{l^e_i}] &= W_{loc} \Tilde{f_t}. \\
\end{split}
\label{eq:prediction}
\end{equation}
$0\le i \le |C|$. $z_i$ are logits generated by a fully connected layer whose parameter is  $W_{nav}$. $a_{c_0}$ indicates the stop action. Thus the action space $|\mathcal{A}| = |C| + 1$ is varied depending on the dynamically built graph.

\begin{figure*}[t]
\centering
\begin{minipage}{.33\textwidth}
    \centering
    \includegraphics[width=0.9\textwidth]{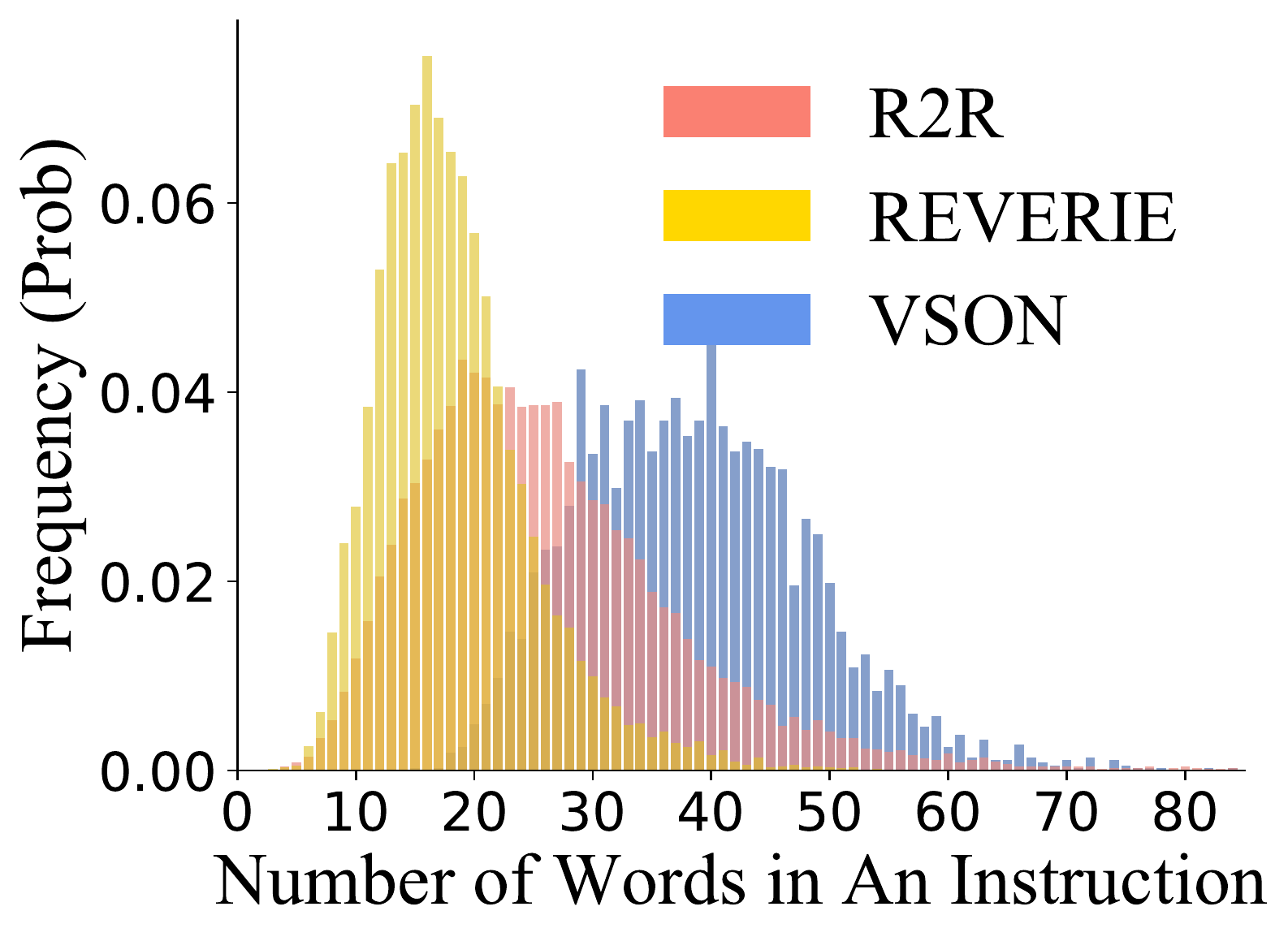}
\end{minipage}
\begin{minipage}{.33\textwidth}
    \centering
    \includegraphics[width=0.9\textwidth]{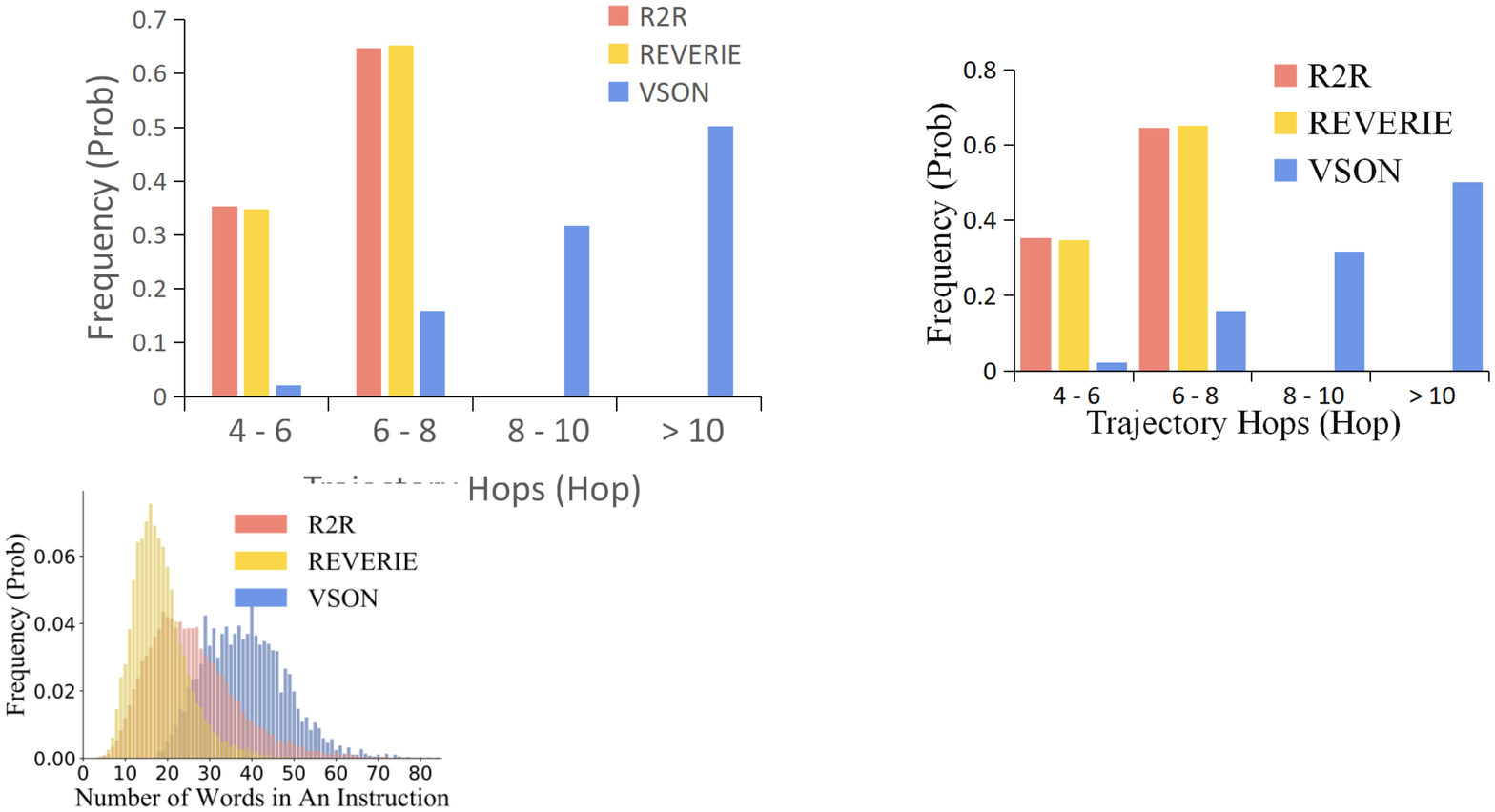}
\end{minipage}
\begin{minipage}{.33\textwidth}
    \centering
    \includegraphics[width=0.9\textwidth]{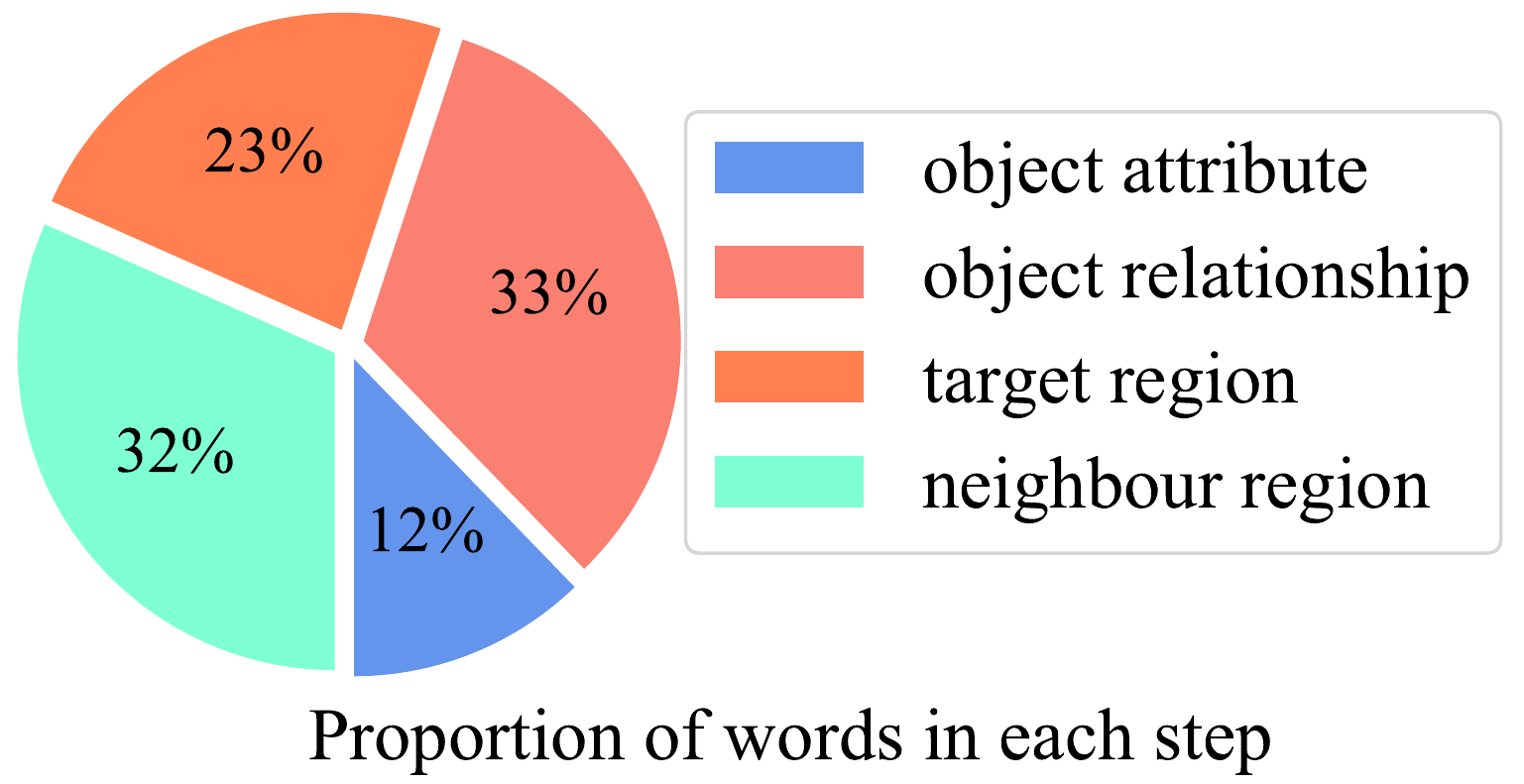}
\end{minipage}
\caption{Statistical analysis across FAO}
\vspace{-9pt}
\label{fig:analysis}
\end{figure*}

\noindent\textbf{Graph-based Exploration} 
Seq2seq navigation models such as speaker-follower~\cite{fried2018speaker} only perceives the current observation and an encoding of the historical information. And existing exploration methods focus on data augmentation~\cite{tan2019learning}, heuristic-aided approach~\cite{ma2019the} and auxiliary task~\cite{zhu2019vision}. 
However, with the dynamically built semantic graph, the navigation agent is able to memorize all the nodes that it observes but has not visited. 
Thus we propose to use the semantic graph to facilitate exploration. 

As shown in Fig.~\ref{fig:model} (yellow box), the graph planner builds the navigation semantic graph during exploration. 
In imitation learning, the navigation agent uses the ground truth action $a^*_t$ to sample the trajectory. 
However, in each step $t$, in graph-based exploration, the navigation action $a_t$ is sampled from the predicted probability distribution of the candidates in Eq.~\ref{eq:prediction}. 
The graph planner calculate the Dijkstra distance from each candidate to the target. 
The teacher action $\hat{a}_t$ is to reach the candidate which is the closest to the target. 
Each trajectory in Room-to-room (R2R) dataset has only one target position. 
However, in the SOON task, since the target object could be able to be observed from multiple positions, trajectories could have multiple target positions. The teacher action $\hat{a}$ is calculated by:
\begin{equation}
\vspace{-1.0 em}
\hat{a_t} = \underset{a_t^{n_i}}{\mathrm{argmin}} \left [ \mathrm{min}\left (   \mathrm{D}(c_i, n_{T_1}),...,\mathrm{D}(c_i, n_{T_m}) \right ) \right ], 
\end{equation}
where $n_{T_1},...,n_{T_m}$ are indexes of $m$ targets, and the action from current position to node $n_i$ is defined by $a_t^{n_i}$. $\mathrm{D}(n_i, n_j)$ stands for the function that calculates the Dijkstra distance between node $n_i$ and $n_j$. Note that the target positions are visible in training to calculate the teacher action but not visible in testing. If the current position is one of target nodes, the teacher actions $\hat{a_t}$ is a stop action. 
Sampling and executing action $a$ from imperfect navigation policy enables the agent to explore in the room. 
Using the optimal action $\hat{a_t}$ helps to learn a robust policy. 


\noindent\textbf{Training Objectives} 
We here introduce two objectives in training: i) the navigation objective $L_{nav}$; ii) the object localization objective $L_{loc}$. The GBE model is jointly optimized by these two objectives. 
In imitation learning, our navigation agent learns from the ground truth action $a^*$. 
In reinforcement learning, the agent learns to navigate by maximizing the discounted reward when taking action $a_t$~\cite{sutton1999policy}. 
In graph-based exploration, we calculate the candidate which is closest to the target by the graph planner and set the action to move to the candidate as $\hat{a_t}$. 
The $L_{nav}$ is the combination of the above three learning approaches: 
\begin{equation}
\begin{split}
& L_{nav} =  - \lambda_{1} \sum_{\tau_1} \sum_t a_t^* \textnormal{log}(p_t) \\
& - \lambda_{2} \sum_{\tau_2} \sum_t a_t \textnormal{log}(p_t) A_t 
 - \lambda_{3} \sum_{\tau_3} \sum_t \hat{a_t} \textnormal{log}(p_t). 
\end{split}
\end{equation}
$A_t$ is the advantage defined in A2C~\cite{mnih2016asynchronous}. The reward of reinforcement learning is calculated by the Dijkstra distance between the current position and the target. 
The $\lambda_1$, $\lambda_2$, $\lambda_3$ are loss weights for imitation learning, reinforcement learning and graph-based exploration respectively. 
Our agent learns a localization branch that is supervised by the center position of the target. Since we map the 2D bounding box position into polar representation, the label consists of two linear values, namely heading $l^h$ and elevation $l^e$. We use Mean Square Error (MSE) to optimize predictions: 
\begin{equation}
L_{loc} = \frac{1}{N} \sum_{i=1}^{N} \left [ (\hat{l^h_i} - l^h_i)^2 + (\hat{l^e_i} - l^e_i)^2 \right ].
\end{equation}
\section{Experiments}

\begin{table*}[t]
\small
 \begin{center}
 \resizebox{1.0\textwidth}{!}{
 {\renewcommand{\arraystretch}{1}
 \setlength\tabcolsep{10pt}
  \begin{tabular}{| l | c c c c | c c c c |}
  
  \hline
    Splits
    & \multicolumn{4}{c|}{Unseen House (Val)}
    & \multicolumn{4}{c|}{Unseen House (Test)}\\
  \hline
  Metrics
  & NE~$\downarrow$ & OSR~$\uparrow$ & SR~$\uparrow$ & SPL~$\uparrow$
  & NE~$\downarrow$ & OSR~$\uparrow$ & SR~$\uparrow$ & SPL~$\uparrow$ \\
 \hline
 Seq2Seq~\cite{anderson2018vision} & 7.81 & 28.4 & 21.8 & - & 7.85 &  26.6 &  20.4 & - \\
 Ghost~\cite{anderson2019chasing} & 7.20 & 44 & 35 & 31 & 7.83 & 42 & 33 & 30\\
 Speaker-Follower~\cite{fried2018speaker} & 6.62 & 43.1 & 34.5 & - & 6.62 & 44.5 & 35.1 & -\\
 RCM~\cite{wang2018reinforced} & 5.88 &  51.9 & 42.5 & - & 6.12 & 49.5 & 43.0 & 38 \\
 Monitor*~\cite{ma2019self} & 5.52 & 56 & 45 & 32 &  5.67 & 59 & 48 & 35\\
 Regretful*~\cite{ma2019the} & 5.32 & 59 & 50 & 41 & 5.69 & 56 & 48 & 40 \\
  EGP~\cite{deng2020evolving} & 5.34 & 65 & 52 & 41 & - & - & - & - \\
  EGP*~\cite{deng2020evolving} & \textbf{4.83} & 64 & \textbf{56} & \textbf{44} & 5.34 & 61 & 53 & 42 \\
  \hline
 GBE (Ours) & 5.20 & \textbf{67.0}
 & 53.9 & 43.4 & \textbf{5.18} & \textbf{64.1} & \textbf{53.0} & \textbf{43.4}\\
  \hline
 \end{tabular}
 }
 }
 \end{center}
 \vspace{-10pt}
 \caption{ The results of the GMSE and previous state-of-the-art methods on R2R (*: model uses additional synthetic data). }
 \vspace{-10pt}
 \label{table:result_r2r}
 \end{table*}

\subsection{From Anywhere to Object (FAO) Dataset}
We provide 3,848 sets of natural language instructions, describing the absolute location in a 3D environment. 
We further collect 6,326 bounding boxes for 3,923 objects across 90 Matterport scenes. Despite the fact that our task does not place limitations on the agent's starting position, we provide over 30K long distance trajectories in our dataset to validate the effectiveness of our task. 
Each instruction contains attributes, relationships and region descriptions to filter out the unique target object when there are multiple objects. 
Please refer to the supplementary materials for more details of our FAO dataset and experimental analysis. 

\noindent\textbf{Data Split}
The training split contains 3,085 sets of instructions with 28,015 trajectories over 38 houses. 
We propose a new split named \emph{validation on seen instruction}, which is a validation set containing the same instructions in the same house with different starting positions. 
The validation seen instruction set contains 245 instructions with 1,225 trajectories. 
The validation set for seen houses with different instructions contains 195 instructions with 1,950 trajectories. 
The validation set for the unseen houses contains 205 instructions with 2,040 trajectories. 

\noindent\textbf{Data Collection}
We first label bounding boxes for objects in panoramic views. Then we convert the bounding box labels into polar representations as described in Sec.~\ref{sec:polar}. Note that the object can be reached from multiple positions. 
We annotate all these positions to reduce the dataset bias. 

To collect diverse instructions with their hierarchical descriptions, we divide the language annotation task into five subtasks as shown in Fig.~\ref{fig:instruction}: 1) Describe the attributes, such as the color, size or shape, of the target; 2) Find at least two objects related to the target and describe their relationship; 3) Conduct explorations in the simulator to describe the region in which the target is located; 4) Explore and describe the nearby regions; 5) Rewrite all descriptions within three sentences. The first four steps ensure language complexity and diversity. 
And the rewriting step makes the language instruction coherent and natural. 

Finally, we generate long navigation trajectories using the navigation graph of each scene. To make the task sufficiently challenging, we first set a threshold of 18 meters. For each instruction and object pair, we fix the target viewpoint and sample the starting viewpoint. We determine a trajectory as valid if the Dijkstra distance between the two viewpoints exceeds the threshold. In some houses, long trajectories are often difficult to find or may even not exist. Thus, we discount the threshold by a factor of 0.8 after every five sample failures. 

\noindent\textbf{Data Analysis} 
Fig.~\ref{fig:analysis} (left) illustrates the distributions of word numbers in the instructions. 
The FAO dataset contains 3,848 instructions with a vocabulary of 1,649 words. The average number of the words in an instruction set is 38.6, while which in REVERIE is 26.3 and in R2R is 18.3.  
Most of the instructions range from 20 words to 60 words, which ensures the power of representation. Moreover, the variance in instruction length makes the description more diverse. 
The trajectory length ranges from 15 meters to more than 60 meters. 
Compared with R2R and REVERIE that most of the trajectories are within 8 hops, as shown in Fig.~\ref{fig:analysis} (middle), FAO provides much more long-term trajectories, which makes the dataset more challenging. 
Fig.~\ref{fig:analysis} (right) illustrates the proportion of word numbers in the four instruction annotating steps. The more words are in the annotation, the richer information it contains. Therefore, we can infer that the object relationship and nearby regions contain the richest information. An agent should consequently pay more attention to these two parts in order to achieve good performance. 
 
 \begin{table*}[t]
\small
 \begin{center}
 \resizebox{1.0\textwidth}{!}{
 {\renewcommand{\arraystretch}{1}
 \setlength\tabcolsep{5pt}
  \begin{tabular}{|l | c c c | c | c c c | c | c c c | c |}
  
  \hline
    Splits
    & \multicolumn{4}{c|}{Val Seen Instruction} 
    & \multicolumn{4}{c|}{Val Seen House}
    & \multicolumn{4}{c|}{Unseen House (Test)}\\
  \hline
  Metrics
  & OSR & SR & SPL & SFPL
  & OSR & SR & SPL & SFPL
  & OSR & SR & SPL & SFPL\\
 \hline
 Human &  - &  - &  - & -  & - & - & - & - & 91.4 & 90.4 & 59.2 & 51.1 \\
 \hline
 Random & 0.1 & 0.0  
 & 1.5 & 1.4 & 0.4 & 0.1 & 0.0
 & 0.9 & 2.7 & 2.1 & 0.4 & 0.0 \\
 Speaker-Follower~\cite{fried2018speaker} & 97.8  & 97.9
 & 97.7  & 24.5 & 69.4 & 61.2 & 60.4 & \textbf{9.1} & 9.8 & 7.0 & 6.1 & 0.6 \\
 RCM~\cite{wang2018reinforced} & 89.1 & 84.0 & 82.6 & 10.9 & 72.7 & 62.4 & 60.9 & 7.8 & 12.4 & 7.4 & 6.2 & 0.7 \\
  AuxRN~\cite{zhu2019vision} & \textbf{98.7} & \textbf{98.4} & 97.4 & 13.7 & \textbf{78.5} & 68.8 & \textbf{67.3} & 8.3 & 11.0 & 8.1 & 6.7 & 0.5 \\
 \hline
 GBE w/o GE & 91.8 & 89.5 & 88.3 & 24.2 & 73 & 62.5 
 & 60.8 & 6.7 & 18.8 & 11.4 & 8.7 & 0.8 \\
  \hline
 GBE (Ours) & 98.6 & \textbf{98.4} &  \textbf{97.9} & \textbf{44.2} & 64.1 & \textbf{76.3}
 & 62.5 & 7.3 & \textbf{19.5} & \textbf{11.9} & \textbf{10.2} & \textbf{1.4} \\
  \hline
 \end{tabular}
 }
 }
 \end{center}
 \vspace{-13pt}
 \caption{The results for baselines and our model on two validation set and test set. }
 \vspace{-0pt}
 \label{table:result}
 \end{table*}
 
\begin{table*}[!t]
    \begin{minipage}{0.49\linewidth}
        \resizebox{\linewidth}{!}{
        \setlength{\tabcolsep}{0.5em}
        {\renewcommand{\arraystretch}{1}
        \begin{tabular}{c|cc|ccc}
        \hline
        \hline
        Models & vision & language & SR & SPL & SFPL \\
        \hline
        GBE & \xmark & \xmark & 0.6 & 0.4 & 0.0 \\
        GBE & \cmark & \xmark & 9.8 & 8.1 & 0.5 \\
        GBE & \xmark & \cmark & 1.8 & 1.5 & 0.2 \\
        GBE & \cmark & \cmark & \textbf{11.9} & \textbf{10.2} & \textbf{1.4} \\
        \hline
        \end{tabular}
        }
        }
        \vspace{-5pt}
        \caption{Ablation of unimodal inputs.}
        \label{tab:abla_input}
    \end{minipage}%
    \hfill
    \begin{minipage}{0.49\linewidth}
        \resizebox{\linewidth}{!}{
        \setlength{\tabcolsep}{1.em}
        {\renewcommand{\arraystretch}{1}
        \begin{tabular}{c|ccc}
        \hline
        \hline
        Models & SR & SPL & SFPL \\
        \hline
        GBE+\circled{1} & 7.3 & 6.2 & 0.5 \\
        GBE+\circled{1}+\circled{2} & 6.2 & 4.9 & 0.7 \\
        GBE+\circled{1}+\circled{2}+\circled{3} & 6.6 & 5.5 & 0.8 \\
        GBE+\circled{4} & \textbf{11.9} & \textbf{10.2} & \textbf{1.4} \\
        \hline
        \end{tabular}
        }
        }
        \vspace{-5pt}
        \caption{Ablation of granularity levels.}
        \label{tab:abla_granularity}
    \end{minipage} 
    \vspace{-10pt}
\end{table*}

\subsection{Experimental Results}

\noindent\textbf{Experiment Setup}
We evaluate the GBE model on R2R and FAO datasets. 
We split our dataset into five components: 1) training; 2) validation on seen instructions (on seen houses as well); 3) validation on seen houses but unseen instructions; 4) validation on unseen houses; and 5) testing. Compared with standard VLN benchmark~\cite{anderson2018vision}, we add a new validation set in FAO, the validation on seen instructions, due to the task starting-independent. 

We evaluate the performance from two aspects: navigation performance and localization performance. The navigation performance is evaluated via commonly used VLN metrics, including Navigation Error (NE), Success Rate (SR), Oracle Success Rate (OSR) and the Success Rate weighted by Path Length (SPL)~\cite{anderson2018on}. 
The localization performance is evaluated by the success rate indicating whether the predicted direction is located in the bounding box. We combine the SPL and localization success to propose a success rate of finding weighted by path length (SFPL): 
\begin{equation}
    \textnormal{SFPL} = \frac{1}{N} \sum_{i=1}^{N} S^{nav}_i S^{loc}_i \frac{l^{nav}_i}{\textnormal{max}(l^{nav}_i, l^{gt}_i)}, 
\end{equation}
where $S^{nav}_i$ and $S^{loc}_i$ are indicators of whether the agent has successfully navigated to or localized the target, respectively. $l^{nav}_i$ is the length of the navigation trajectory, while $l^{gt}_i$ is the shortest distance between the ground truth target and the starting position. 

\noindent\textbf{Implementation Details}
We compare the proposed model with several baselines: 1) a random policy; 
2) Speaker-Follower~\cite{fried2018speaker}, an imitation learning method; 
3) RCM~\cite{wang2018reinforced}, an imitation learning and reinforcement learning; 
4) AuxRN~\cite{zhu2019vision}, a model with auxiliary tasks; 
5) the Hierarchical Memory Network. 
All five models employ the same vision language navigation backbone introduced in Sec.~\ref{sec:GBE}. 
The visual encoder $g$ is implemented by a Resnet-101~\cite{he2016deep} and the language encoder $h$ is a combination of a word embedding layer and an LSTM~\cite{hochreiter1997long} layer. 
We train all models on the training split for 10K interactions to ensure that all models are sufficiently trained. The optimizer we use is RMSProp and the learning rate is $10^{-4}$. 

\noindent\textbf{Results on R2R}
In Tab.~\ref{table:result_r2r}, we compare the GBE model with state-of-the-art models without pretraining and auxiliary tasks. 
On the unseen house validation set, the GBE outperforms all models without using additional data. 
It outperforms EGP, other graph-based navigation method by 2.4\% in SPL. 
On the test set, the GBE outperforms pervious models on all the evaluation metrics. 
It outperforms RCM, a seq2seq model with imitation learning with reinforcement learning by 5.4\% in SPL.

\noindent\textbf{Results on FAO}
The experimental results are presented in Tab.~\ref{table:result}. The performances of the baseline models reveal some unique features of the FAO dataset. 
Firstly, the human performance largely outperforms all models. The existence of this human-machine gap suggests that current methods are not able to solve this new task. 
The random policy method performs poorly on all metrics, which reveals that our dataset is not biased. 
Moreover, Reinforced Cross-Modal Matching (RCM), a method combines imitation learning and reinforcement learning outperforms the pure imitation learning method (Speaker-follower) on the unseen house set. 
It indicates that reinforcement learning helps avoid overfitting in our dataset. 
Our experiment of the AuxRN shows that the auxiliary tasks work on R2R are not benefitial on FAO, which indicate the SOON is unique. 
We test the performance of the GBE and the GBE without graph-based exploration. 
We observe that with graph-exploration, the model obtain better generalization ability. The final model is 0.7\% higher in oracle success rate, 0.5\% higher in success rate, 1.5\% higher in SPL and 0.6\% higher in SFPL than which without graph-based exploration on the test set. 
We discover that models perform well on the seen instruction set but perform poorly on other two sets. 
Since the domain of the seen instruction set is close to the training set, it indicates that models fit the training data well but lack of generalizability. 

\noindent\textbf{Ablation study of FAO}
We ablate the FAO dataset from two aspects: 1) the effect of vision and language modalities and 2) the effect of different granularity levels. 
The ablation result of input modal is shown in Tab.~\ref{tab:abla_input}. 
We observe that the model without vision and language input performs the worst. 
Thus it is impossible to finish SOON task without vision-language modalities. And the model with vision only performs better than the model with language only. 
We infer that the vision is more import than language in SOON task. Finally, we find that the model with vision and language performs the best, indicating that the two modalities are related and both modalities are important. 
Some objects like `chair' exist in all houses while other objects like `flower' do not commonly exist.  The model learns prior knowledge to find common object in navigation without language.

The ablation result of granularity levels is shown in Tab.~\ref{tab:abla_granularity}. 
We train the GBE with different annotation granularity levels: \circled{1} object names, \circled{2} object attributes and relationships, \circled{3} region information, \circled{4} rewritten instructions. 
Note that the model with object names (GBE+\circled{1}) is equivalent to the \emph{ObjectGoal} navigation. 
We find that the model trained in \emph{ObjectGoal} setting performs worse than the models trained with more information. It has two reasons: 1) there are more than one objects belongs to the same class, and navigating with object name cause ambiguity; 2) navigating without scene and region makes the agent harder to find the final location. 
By comparing the first three experiments, we infer that the object name (\circled{1}), object attributes and relationships (\circled{2}) and region descriptions (\circled{3}) all contribute to the SOON navigation. 
At last, we find that the model with rewritten instructions performs the best (0.6\% higher in SFPL than GBE+\circled{1}+\circled{2}+\circled{3}). We infer that a well developed natural language instruction facilitates the agent to comprehend. 

\section{Conclusion}

In this paper, we have proposed a task named Scenario Oriented Object Navigation (SOON), in which an agent is instructed to find an object in a house from an arbitrary starting position. To accompany this, we have constructed a dataset named From Anywhere to Object (FAO) with 3K descriptive natural language instructions.
To suggest a promising direction for approaching this task, we propose GBE, a model that explicitly models the explored areas as a feature graph, and introduces graph-based exploration approach to obtain a robust policy. Our model outperforms all previous state-of-the-art models on R2R and FAO datasets. We hope that the SOON task could help the community approach real-world navigation problems. 

\section{Acknowledgements}
This work was supported in part by  National Key R\&D Program of China under Grant No. 2020AAA0109700,  Natural Science Foundation of China (NSFC) under Grant No.U19A2073, No.61976233 and No.61906109, Guangdong Province Basic and Applied Basic Research (Regional Joint Fund-Key) Grant
No.2019B1515120039,  Shenzhen Outstanding Youth Research Project (Project No. RCYX20200714114642083) Shenzhen Basic Research Project (Project No. JCYJ20190807154211365), Zhijiang Lab’s Open
Fund (No. 2020AA3AB14) and CSIG Young Fellow Support Fund. 
And by the Australian Research Council Discovery Early Career Researcher Award
(DE190100626).



\end{document}